\documentclass[10pt,twocolumn,letterpaper]{article}

\usepackage[pagenumbers]{cvpr} 

\usepackage{algorithm}
\usepackage{algpseudocode}
\usepackage{amsmath}
\usepackage[T1]{fontenc}

\definecolor{cvprblue}{rgb}{0.21,0.49,0.74}
\usepackage[pagebackref,breaklinks,colorlinks,allcolors=cvprblue]{hyperref}

\def\paperID{8}
\def\confName{CVPR}
\def\confYear{2025}

\title{Efficient Self-Supervised Learning for Earth Observation via Dynamic Dataset Curation}

\author{Thomas Kerdreux$^{1,*}$ \quad Alexandre Tuel$^{1}$ \quad Quentin Febvre$^{2}$ \quad Alexis Mouche$^{2}$ \\ \quad Bertrand Chapron$^{2}$ \vspace{0.3em} \\
{\normalsize $^1$Galeio, Paris, France} \quad
{\normalsize $^2$Ifremer, UMR CNRS LOPS, Brest, France} \quad \\
{\normalsize $^*$Corresponding author: \href{mailto:tkerdreux@galeio.fr}{tkerdreux@galeio.fr}}
}

\begin{document}
\maketitle

\begin{abstract}
Self-supervised learning (SSL) has enabled the development of vision foundation models for Earth Observation (EO), demonstrating strong transferability across diverse remote sensing tasks. While much research has focused on network architectures and training strategies, the role of dataset curation -- particularly in balancing and diversifying pre-training datasets -- remains underexplored. In EO, this challenge is exacerbated by the strong redundancy and heavy-tailed distributions of satellite imagery, which can lead to biased representations and inefficient training.\\
In this work, we introduce a dynamic dataset pruning strategy designed to enhance SSL pre-training efficiency by maximizing dataset diversity and balancedness. Our method iteratively refines the training set without relying on a pre-existing feature extractor, making it well-suited for domains where curated datasets are unavailable. We illustrate our approach on the Sentinel-1 Wave Mode (WV) Synthetic Aperture Radar (SAR) archive, a challenging dataset primarily composed of ocean observations. We train models from scratch on the entire Sentinel-1 WV data archive over 10 years. Our results, validated across three downstream tasks, show that dynamic pruning improves both computational efficiency and feature quality, leading to better transferability in real-world applications. This work provides a scalable and adaptable solution for dataset curation in EO, paving the way for more efficient and generalizable foundation models in remote sensing.\\
We release the weights of OceanSAR-1, the first foundation model in our OceanSAR models family — a series of models dedicated to ocean observation and analysis using SAR imagery, at \href{https://github.com/galeio-research/OceanSAR-models/}{github.com/galeio-research/OceanSAR-models/}.

\end{abstract}

\section{Introduction}
\label{sec:intro}

Self-supervised learning (SSL) has recently demonstrated much potential for Earth Observation (EO). Vision foundation models, in particular, have shown a remarkable ability to learn task-agnostic image representations without requiring any human labels. A wide range of EO foundation models are now available, covering multiple modalities, whether optical, radar, multispectral or hyperspectral data \cite[e.g.,][]{braham2024,Guo_2024_CVPR,hong2024spectralgpt,marsocci2024pangaea,li2025sar,lu2025visionfoundationmodelsremote}. Much effort has been put into exploring new network architectures and training strategies to improve the performance of these foundation models, and in particular their ability to handle multiple remote sensing modalities. However, one aspect of model pre-training has so far attracted relatively little attention, namely the selection of images to build pre-training datasets.

There is now a large body of evidence that SSL model performance on downstream prediction tasks is strongly dependent on the scale, on the diversity, and on the balancedness of the pre-training datasets \cite{goyal2021self,balestriero2023cookbook,oquab2023dinov2,vo2024automaticdatacurationselfsupervised}. By contrast, training vision models on random collections of images leads to a drop in performance \cite{tian2021dividecontrastselfsupervisedlearning}, likely because of the heavy-tailed distribution of concepts in uncurated datasets \cite{liu2019largescalelongtailedrecognitionopen}.
Large Language Models (LLMs) are trained on carefully curated, high-quality datasets \cite{touvron2023llama,naveed2024llm}. The challenge of dataset balancedness is particularly acute in EO because remote sensing imagery exhibits a highly non-uniform distribution, with some concepts, like oceans, deserts, or tropical forests, being much more frequent than others. Such a strong imbalance may lead to biases toward a few dominant scenes in the learned representations, as well as a waste of computational resources during training.

To overcome this challenge, most existing EO foundation models rely on a handful of pre-compiled, sometimes carefully curated datasets, like BigEarthNet \cite{Sumbul_2019}, SSL4EO \cite{wang2023ssl4eo}, SAtlasPretrain \cite{bastani2023satlaspretrain} or FMoW \cite{christie2018functionalmapworld}. Data curation in these datasets is typically achieved by selecting images according to a criterion, often based on land cover, meant to ensure balancedness in the resulting dataset \cite{Sumbul_2019,sumbul2021bigearthnetdatasetnewclassnomenclature,wang2023ssl4eo}. A similar approach was taken in the Prithvi model \cite{szwarcman2025prithviv2}, whereby individual Harmonized Landsat/Sentinel-2 images were selected based on average climatology and land cover to build the model's pre-training dataset.

However, such approaches have important drawbacks. First, they require an a priori selection criterion to build the pre-training dataset. While land cover may certainly make sense for land-based imagery, it is not the only choice one could think of. Besides, for some modalities, like SAR imagery over the ocean, devising such a criterion may require substantial efforts that are not replicable for other modalities (for instance, diversity in ocean imagery is largely driven by changing weather conditions). Furthermore, there is no guarantee that the resulting datasets are truly balanced, since the selection criteria, like land cover, are always reductive (diversity in land imagery is not only driven by differences in land cover, but also by seasonality, weather, extreme events, etc.)

We address this challenge by introducing a simple, yet effective strategy to automatically prune the pre-training dataset in SSL to maximize its diversity and balancedness. To do so, we build on previous work \cite{vo2024automaticdatacurationselfsupervised} to define a dynamic coreset selection strategy over time for highly redundant SSL datasets, particularly in scenarios where a reliable pre-existing feature extractor is unavailable for pruning.
We apply our method to radar imagery over the oceans, a choice example of a \textit{strongly redundant} EO dataset. Using three different downstream tasks, we show how dynamic pruning increases model performance while improving computational efficiency. Our results demonstrate that selecting a diverse and balanced subset of pre-training data leads to improved feature representations, ultimately enhancing the transferability of self-supervised models to real-world applications. By validating our approach across multiple tasks, we highlight its robustness and generalizability, offering a practical solution for curating large-scale, redundant EO datasets in the absence of predefined feature extractors. We also show that a carefully-trained, single-modality foundation model can outperform large, multi-modal models.

\section{Related Work}

\paragraph{Dataset pruning/coreset selection}
Scaling laws highlight the balance between dataset size, diversity, and model size as key factors in achieving optimal performance in self-supervised learning. These empirical laws suggest that performance improvements require exponentially increasing dataset sizes. However, it is known that naive data pruning can be a promising approach to mitigate the constraints imposed by these scaling laws \cite{sorscher2022beyond}.

In practice, self-supervised training across different domains, including images \cite{birodkar2019semantic}, language \cite{geiping2023cramming,albalak2024survey}, and vision-language models \cite{nguyen2022quality,wang2023too,cao2023moreremovingtextregionsimproves}, commonly involves an initial curation phase of the large datasets. This curation process often goes beyond simple deduplication \cite{tirumala2024d4} and aims to remove semantic redundancies within the dataset.

Numerous dataset pruning, coreset selection, and dataset distillation strategies have been developed within machine learning and deep learning to influence model training by effectively modifying the training dataset. These strategies follow various heuristics and objectives, such as leveraging model optimization status to encourage learning from harder examples—such as focal loss or self-paced reweighting \cite{vito2022asymmetric,ding2022enhancing,hou2023improving}; progressively increasing data sample difficulty using augmentation techniques \cite{guo2024improving,ye2021efficient,chu2021cuco};  utilizing label information for pruning \cite{sener2018activelearningconvolutionalneural,mirzasoleiman2020coresets,xia2023moderate}; or pruning to ensure better data coverage \cite{zheng2023coveragecentriccoresetselectionhigh,yang2024mind}.

The curation of large datasets for self-supervised training presents distinct challenges, particularly in cases where no effective feature extractor exists or where datasets exhibit high redundancy. For EO datasets, most existing approaches rely on metadata to guide dataset sampling, such as balancing datasets based on land cover types or urbanization patterns. However, in cases where metadata are unavailable or insufficiently informative, alternative/complementary strategies must be explored to ensure dataset balance and effective training outcomes.

Recent work has sought to perform unsupervised dataset curation via hierarchical clustering in the latent space \cite{vo2024automaticdatacurationselfsupervised}. However, the proposed approach still requires an existing feature extractor and cannot therefore be considered fully unsupervised. We build on these efforts to develop a dynamic sampling strategy that functions even without a pre-existing feature extractor. As a case study, we focus on a niche modality -- Sentinel-1 Synthetic Aperture Radar (SAR) imagery in wave mode -- to demonstrate the effectiveness of this approach. In this example, the training dataset can be considered overly redundant, as wave mode data mainly consists of images of the ocean, which are exceedingly similar.

\paragraph{Sentinel-1 SAR foundation models} Multiple SAR foundation models have already been published in the literature.
\begin{itemize}
    \item CROMA (Contrastive Radar-Optical Masked Autoencoders) \cite{fuller2023croma}: CROMA is based on a hybrid training strategy mixing MAE-type reconstruction with a contrastive (InfoNCE) loss, using optical and radar images as positive pairs. CROMA was pre-trained on the SSL4EO-S12 \cite{wang2023ssl4eo} dataset -- a geographically and seasonally diverse dataset of 1 million paired Sentinel-2 L2A and Sentinel-1 IW GRD samples.

    \item DeCUR (Decoupling Common and Unique Representations) \cite{wang2024decoupling}: the idea behind the  approach is to take into account the fact that, in multi-modal self-supervised learning, some modalities usually contain information that is not present in other modalities (for example, radar sensors will see through clouds, but not optical ones \cite{Gawlikowski2024}). DeCUR therefore relies on cross-correlation matrices between embeddings from different modalities to separate between cross-modal common ones and modality-unique ones. Intra-modal representation are also enhanced using a regularization term on modality-specific embeddings to avoid collapse. The model was also pre-trained on the SSL4EO-S12 \cite{wang2023ssl4eo} dataset.

    \item MoCo (Momentum Contrast) \cite{wang2023ssl4eo}: like DINO \cite{goyal2021self}, the method we use in our experiments, MoCo is a contrastive self-supervised learning strategy that involves building dynamic dictionaries of embeddings. In the InfoNCE loss, instead of comparing an image to the ones from the same batch, MoCo allows to query from a larger queue, in which embeddings are computed on-the-fly using a momentum-updated encoder. The model was pre-trained on the SSL4EO-S12 \cite{wang2023ssl4eo} dataset.

    \item DOFA (Dynamic One-For-All) \cite{xiong2024neural}: DOFA avoids using separate encoders when training on multiple modalities by using a dynamic weight generator network that adjusts the weights of a shared vision backbone according to the wavelength(s) of the spectral bands of the input image. The model was pre-trained on a variety of datasets, including Sentinel-1 IW GRD images from the SAtlasPretrain \cite{bastani2023satlaspretrain} dataset.

    \item SoftCon \cite{wang2024multi}: unlike other strategies, the SoftCon (Soft Contrastive learning) strategy relies on automatically generated image labels to compute a modified InfoNCE loss, which weighs the similarity between images in the same batch according to their labels. Labels are obtained by matching data from Google's Dynamic World land cover maps with images from the SSL4EO-S12 \cite{wang2023ssl4eo} dataset, which is used for pre-training. The training also involves Siamese masking of image patches for enhanced robustness.

    \item WV-net \cite{glaser2024wv}: a ResNet50 model trained on Sentinel-1 WV images using a SimCLR strategy, along with a set of complex data augmentation techniques designed specifically for SAR imagery. WV-net was pre-trained on the entire Sentinel-1 WV archive, at the time consisting of about 9.9 million images.
\end{itemize}

All these models (except for WV-net) were pre-trained on Sentinel-1 IW GRD only, which consists of intensity-only, dual polarization (usually VV+VH) SAR images. WV images, by contrast, are only acquired with the VV polarization.

\section{Methodology and Results}

\subsection{Sentinel-1 WV data}

Our work is based on Synthetic Aperture Radar (SAR) data acquired by the Sentinel-1 constellation, specifically \textit{WaVe mode} (WV) data acquired over the oceans. The ESA Sentinel-1 mission is a constellation of two polar-orbiting, sun-synchronous satellites (S-1 A and S-1 B) launched in April of 2014 and 2016, respectively. They are equipped with a C-band SAR instrument with frequency 5.405 GHz (5.5 cm wavelength). They have a 12-day repeat cycle at the equator, and are phased at 180$^{\circ}$ to provide an effective 6-day repeat cycle. S-1 B ceased to function in late 2021, while S1-C was launched in December 2024 and is still in commissionning phase. The launch of S1-D is planned for late 2025. These satellites operate in four different acquisition modes: WV, Interferometric Wide (IW), Extra Wide (EW) swath and Stripmap (SM) modes. These modes represent different acquisition strategies (chiefly in terms of swath width, image footprint, resolution and polarisations), and are deployed over different geographies. WV data is acquired almost exclusively over oceans (a few images cover Africa, Australia and the Great Lakes region of North America). It consists in $\approx$ 20$\times$20 km images with a spatial resolution of 5$\times$5 m, every 100 km along the orbit, acquired alternately on two different incidence angles ($\approx$23$^{\circ}$ and 36$^{\circ}$). WV data are only available in single-look complex (SLC) format, while the more generally used IW data are also available in ground range detected (GRD) format.\\
The WV data archive we used includes about 12 million different images, covering the full 2015-2024 period (each satellite produces approximately 60,000 WV images per month). Their spatial distribution is very uneven. Some regions, such as the South Pacific Ocean, have a much higher concentration of images, while others, notably the North Atlantic, are more sparsely covered (European waters are observed in IW and EW modes). To build an initial training database, we therefore sample from these 12 million images uniformly in space and time, retaining about 2 million different images. We make sure during this step to exclude all images used in our benchmarks (see section \ref{downstream_tasks}). To reduce the computational load, we downsample images to a 50-meter resolution, which preserves sufficient information richness for model training and relevant benchmarking.\\
Sentinel-1 WV images are characterized by a high level of redundancy. Many images exclusively display ocean waves, and even when other phenomena are present (sea-ice or rain cells, for instance), the background still chiefly consists of ocean waves. As a consequence, many phenomena that can be seen in WV images are rare, relatively speaking, compared to ocean waves. This results in significant unbalancedness in the dataset.

\subsection{Training Framework}

For our experiments, we employ the DINO self-supervised learning strategy \cite{caron2021emerging}. DINO is a contrastive learning strategy that relies on self-distillation between a teacher and a student model. It consists in learning a global representation of an image using a mixture of global and local views of that image. Global views pass through the teacher model, which is tasked with predicting high-level features, while the student model is tasked with predicting these same features from the local views. Regarding network architectures, we use Vision Transformers (ViTs), specifically ViT-Small and ViT-Base, with respectively 27 and 82 million parameters. For a fair comparison with some of the existing SAR foundation models, we also train a ResNet50 in the same conditions. This choice aligns with the unimodal nature and high redundancy of the dataset, where excessively large models would probably be unnecessary. Note, however, that our proposed approach is compatible with any training strategy and network architecture.\\

\subsection{Dynamic coreset selection for highly redundant dataset}

We present our simple dynamic sampling strategy in Algorithm \ref{alg:ssl_redundant_training}. We start by training a model with the selected SSL loss (in this case, the DINO loss) for a pre-specified number of warmup epochs, until a sufficiently effective feature extractor is learned. Then, at scheduled intervals during training, the original dataset is clustered, and a subset of it is sampled. The SSL model then continues training for the remaining epochs until the next resampling step on this dynamically selected data.

\begin{algorithm}[h!]
    \caption{SSL on an Overly Redundant Dataset (SSL-ORD)}
    \label{alg:ssl_redundant_training}
    \begin{algorithmic}[1]
        \Require Dataset $\mathcal{D}$, Total training epochs $T$, Warm-up epochs $T_{\text{w}}$, Sampling epoch frequency $T_{\text{s}}$, SSL training algorithm $\mathcal{A}(\cdot,\cdot,\cdot)$, Pruning strategy $\mathcal{S}(\cdot,\cdot,\cdot,\cdot)$, Reduction ratio $\rho$, Sampling sequence $(\eta_i)$.
        \Ensure Trained model $\theta$

        \State Initialize model parameters $\theta_0$
        \State $\theta_{T_{\text{w}}-1} \leftarrow \mathcal{A}(\theta_0, \mathcal{D}, T_{\text{w}})$  \Comment{Train without dataset pruning}

        \For{epoch $e = T_{\text{w}}$ to $T$}
            \If{$e \mod T_{\text{s}} = 0$}  \Comment{Apply dataset pruning at scheduled intervals}
                \State $\mathcal{D}_{e+1} \leftarrow \mathcal{S}(\theta_e, \mathcal{D}, \rho, \eta_e)$
            \Else
                \State $\mathcal{D}_{e+1} \leftarrow \mathcal{D}_e$
            \EndIf
            \State $\theta_{e+1} \leftarrow \mathcal{A}(\theta_e, \mathcal{D}_e, 1)$
        \EndFor
    \end{algorithmic}
\end{algorithm}

Algorithm \ref{alg:ssl_redundant_training} relies on access to two key procedures. The self-supervised learning (SSL) training algorithm $\mathcal{A}(\cdot, \cdot, \cdot)$ takes a model checkpoint $\theta$ and performs SSL on the dataset $\mathcal{D}$ for $T$ epochs. This approach is therefore highly versatile with respect to SSL methods. While we conducted experiments using DINO for simplicity, any SSL strategy could be employed within this framework. It provides a flexible foundation for conducting extensive numerical experiments across diverse training structures by relying solely on the features extracted from the SSL-trained model. Interestingly, certain SSL methods compute additional latent concepts beyond extracted features, potentially providing valuable insights for improving unsupervised dataset clustering. Exploring a deeper integration between the pruning strategy and SSL mechanisms would be a fruitful direction for future research.

Because this approach does not require a pre-existing feature extractor, it is especially useful in scenarios where none exists -- such as for \textit{de novo} modalities where reliable feature extractors have yet to be developed. This occurs when a balanced and diverse dataset like ImageNet is unavailable or when no self-supervised models have been trained (or publicly released) on the specific modality.

Algorithm \ref{alg:ssl_redundant_training} relies on a pruning strategy $\mathcal{S}(\cdot, \cdot, \cdot, \cdot)$, which, given a (self-supervised) trained feature extractor $\theta$, prunes the dataset $\mathcal{D}$ to retain only a fraction $\rho$ of the original datapoints. The pruning strategy typically consists of two steps: a clustering phase followed by a sampling phase. The sampling parameter $\eta$ controls the evolution of the sampling strategy during training, allowing for dynamic adjustments that may enhance learning. In practice, we experimented with a modified version of \cite[Algorithm 1]{vo2024automaticdatacurationselfsupervised}, as described in Algorithm \ref{alg:hierarchical_kmeans_resampling}. This clustering-based pruning approach was designed to rebalance large training datasets in self-supervised learning, ensuring that the retained subset remains diverse and representative throughout training.

\begin{algorithm}[h!]
    \caption{\textit{Modified} Hierarchical k-Means with Resampling Algorithm \cite{vo2024automaticdatacurationselfsupervised}}
    \label{alg:hierarchical_kmeans_resampling}
    \begin{algorithmic}[1]
        \Require Dataset $\mathcal{D} \in \mathbb{R}^{n \times d}$, Size of dataset subset $n_{\text{c}}$,
        Number of levels $L$, Clusters per level $(k_i)_{1 \leq i \leq L}$, Sampling diversity $\eta\in[0, 1]$, Reduction ratio $\rho$, Hierarchical clustering algorithm \textsc{H-KMeans} \cite[Algorithm 1]{vo2024automaticdatacurationselfsupervised}.
        \Ensure Balanced dataset $\mathcal{D}_{\rho}$ where $\left\lceil \frac{|\mathcal{D}_{\rho}|}{|\mathcal{D}|} \right\rceil = \rho$
        \State $I_{n_{\text{c}}} \gets \text{RandomSubset}(\{1, \dots, n\}, n_{\text{c}})$ \Comment{Randomly sample a subset of indices}
        \State $\mathcal{D}_{\text{small}} \gets \mathcal{D}[I_{n_{\text{c}}}]$ \Comment{Extract subset from dataset}
        \State $(C_t)_{1 \leq t \leq L} \leftarrow \textsc{H-KMeans}(\mathcal{D}_{\text{small}}, (k_i)_{1 \leq i \leq L})$ \Comment{Perform hierarchical clustering}
        \State $L_T \gets \Call{assign}{\mathcal{D}, C_L}$ \Comment{Assign clusters based on final centroids $C_L$}
        \State $\mathcal{D}_{\rho} \gets \Call{resample}{L_T, \rho, \eta}$ \Comment{Resample data to achieve desired reduction}
    \end{algorithmic}
\end{algorithm}

Algorithm \ref{alg:hierarchical_kmeans_resampling} differs from \cite[Algorithm 1]{vo2024automaticdatacurationselfsupervised} in that it does not perform clustering on the entire dataset. Instead, our primary motivation for pruning is to significantly reduce dataset redundancy rather than to capture its underlying concepts. This approach ensures that dynamic dataset sampling incurs no additional computational burden on the GPUs, as the clustering process is lightweight enough to be offloaded to the CPU during training. If pruning is scheduled for epoch $e$, we store the embeddings of the first $N$ data batches, run step 3 of Algorithm \ref{alg:hierarchical_kmeans_resampling} on the CPU, and progressively run steps 4 and 5 as the remaining data batches within epoch $e$ are processed by the GPU.

A second difference is the introduction of the sampling diversity parameter $\eta\in[0,1]$, which controls how datapoints are selected within each cluster. Specifically, $\eta$ determines whether sampling favors points near the centroid (\textit{i.e.}, more representative) or those on the cluster boundary (\textit{i.e.}, less representative). As previously observed \cite{vo2024automaticdatacurationselfsupervised}, empirical results suggest that sampling closer to the centroids performs best when using a single pruned dataset, as this promotes balance and well-discriminated concepts.

\subsection{Experiments}

We run experiments using a ResNet50, a ViT-S/16, a ViT-S/8 and a ViT-B/8\cite{dosovitskiy2020image}. In each case, we pre-train the model on our WV dataset without any dataset pruning, with pruning that discards 50\% of the dataset, and with pruning that discards 75\% of the dataset. We therefore conduct a total of 12 runs. For each run, we train the model for 200 epochs. For the runs that include pruning, we adjust the number of epochs so that the models are trained on the same number of batches as the ones without pruning (with a 20-epoch warmup before the first pruning, this means a total of 380 epochs for the 50\% pruning and 740 epochs for the 75\% pruning). All runs are conducted with a batch size of 256, an AdamW optimizer, a cosine scheduler with annealing, and a 0.0001 initial learning rate.

\subsection{Downstream Tasks}
\label{downstream_tasks}

To quantify the performance of foundation models on Sentinel-1 WV imagery, we use three different labelled datasets, covering two kinds of tasks: classification and regression. All the datasets are randomly split between training (80\%) and validation (20\%).\\
We report the performance of available, SOTA models trained on Sentinel-1 dual-polarization imagery, namely CROMA\cite{fuller2023croma}, MoCo\cite{wang2023ssl4eo}, DeCUR\cite{wang2024decoupling}, DOFA\cite{xiong2024neural}, and SoftCon\cite{wang2024multi}. Note that the WV-Net\cite{glaser2024wv} model is not publicly available, and therefore we did not include it in this study. For MoCo and DeCUR, we use the model versions available from the torchgeo package \cite{Stewart_TorchGeo_Deep_Learning_2022}.

\subsubsection{TenGeoP dataset}

\begin{figure*}
   \centering
   \includegraphics[width=0.9\linewidth]{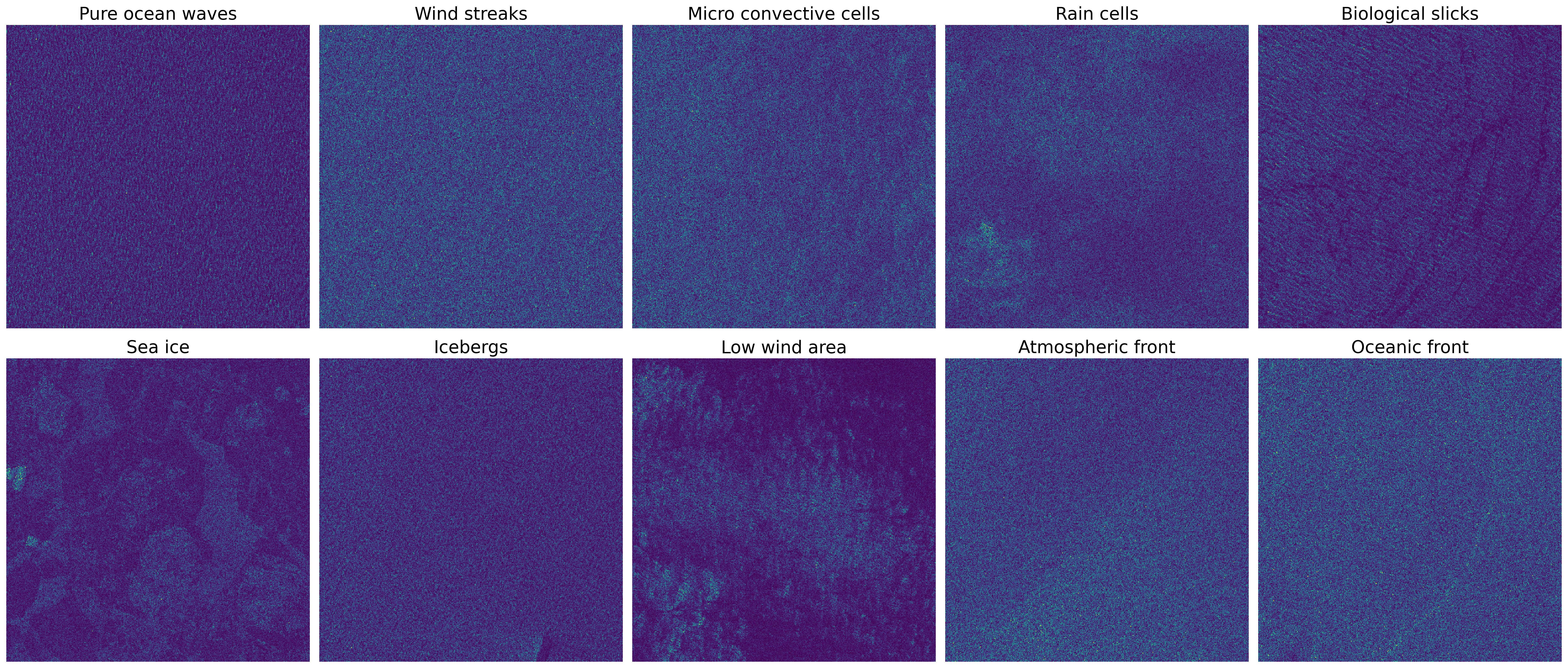}
   \caption{Illustration of the 10 geophysical classes in the TenGeoP dataset.}
   \label{fig:tengeop}
\end{figure*}

The TenGeoP dataset \cite{wang2019labelled} consists of 37,553 WV images that were manually annotated by human SAR experts into ten categories. These correspond to ten different geophysical phenomena, of both oceanic and
meteorologic nature (pure ocean waves, wind streaks, micro convective cells, rain cells, biological slicks, sea ice, icebergs, low wind areas, atmospheric fronts and oceanic fronts). For each image, only one category was selected. The dataset covers the entire ocean and is composed of Sentinel-1A WV images acquired in 2016. Sample images are shown in Figure \ref{fig:tengeop} The dataset is available at \href{https://doi.org/10.17882/56796}{https://doi.org/10.17882/56796}.

\subsubsection{Significant Wave Height dataset}

We randomly selected 10,000 WV images and labeled them with their Significant Wave Height ($H_s$), defined as the mean height of the highest third of the waves passing through a point over a given period. The $H_s$ values were obtained from co-located altimeter measurements available in the CMEMS database \cite{CMEMS}.
WV and altimeter data were considered co-located if acquired within 3 hours of each other and if the altimeter measurement was within $\pm$2$^{\circ}$ of the WV image center. When multiple altimeter measurements met these criteria, we retained the one closest in space. The data will be made publicly available.

\subsubsection{Wind speed dataset}

SAR images capture information on the sea-surface roughness, which is related to surface wind speed \cite{ahsbahs2020windspeed,hauser2023sarwind}. We used a random subset of 50,000 images among those selected by \citet{odriscoll2023obukhov} and labeled with wind speed using ERA5 reanalysis data \cite{hersbach2020era5} as ground truth. The data are available at \href{https://doi.org/10.5281/zenodo.7784019}{https://doi.org/10.5281/zenodo.7784019}.

\subsection{Results}

\subsubsection{Impact of dataset pruning}

Table \ref{tab:impact_pruning} summarizes the results of our experiments and highlights the impact of data pruning on model performance on downstream tasks. We find that pruning the dataset leads to an overall increase in performance on downstream tasks, that can be very significant (for instance from 2.5 to 3\% for the TenGeoP classification task), at no additional computation cost. The increase is smaller, however, for the two regression tasks (on the order of 0.01-0.02 RMSE). A possible reason for this is that DINO may not be an ideal training strategy for such tasks, as it favors a holistic understanding of images and may overlook finer-scale texture details that are important for SWH and wind speed prediction.

\begin{figure}[t]
   \centering
   \includegraphics[width=0.9\linewidth]{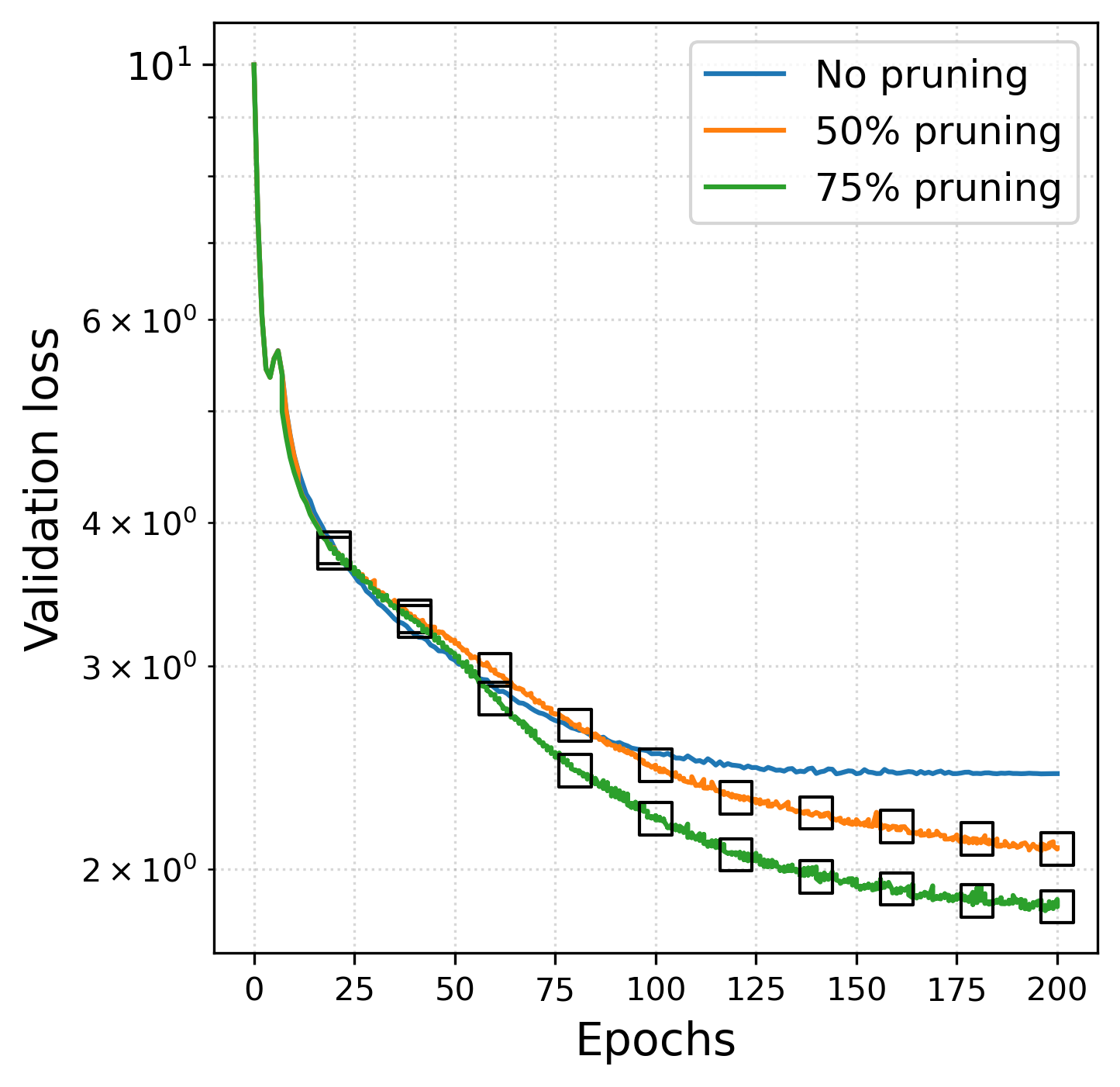}
   \caption{DINO validation loss as a function of training epoch for a ViT-S/16 model, with (blue) no data pruning, (orange) pruning that discards 50\% of the training dataset and (green) pruning that discards 75\% of the training dataset. For the latter two curves, black squares indicate epochs at which dataset pruning is performed.}
   \label{fig:val_loss}
\end{figure}

\begin{figure}[t]
   \centering
   \includegraphics[width=0.9\linewidth]{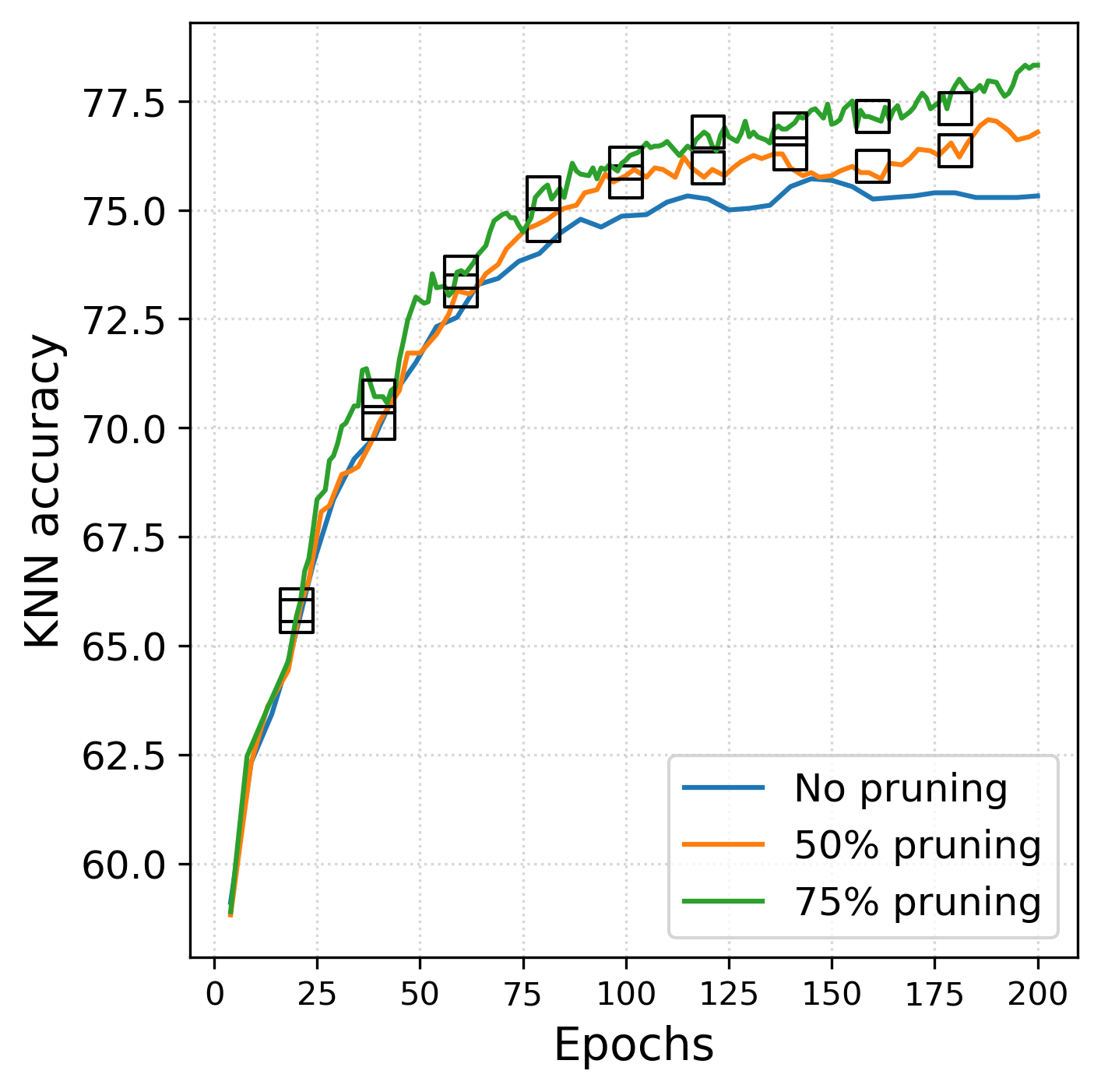}
   \caption{Same as Figure \ref{fig:val_loss}, but for the k-NN classification accuracy on TenGeoP.}
   \label{fig:knn_accuracy}
\end{figure}

Furthermore, validation loss and k-NN probing curves during training suggest that experiments with pruning would benefit from even more training, since model performance does not plateau when reaching our limit of 200 epochs, which is not the case for the no-pruning experiments (Figures \ref{fig:val_loss} and \ref{fig:knn_accuracy}). Since images from downstream datasets were excluded from the pre-training dataset, this indicates better and stronger generalizability by the with-pruning models, since they achieve a higher accuracy while being (effectively) trained on a smaller set of data.

\begin{table}[h!]
  \centering
  \begin{tabular}{@{}lcccccc@{}}
    \toprule
    Model & Pruning & TenGeoP & SWH & Wind speed \\
    \midrule
    ResNet50 & 0\% & 72.7 & 0.72 & 1.43 \\
    ResNet50 & 50\% & 73.8 & 0.71 & 1.42 \\
    ResNet50 & 75\% & 75.5 & 0.70 & 1.42 \\
    ViT-S/16 & 0\% & 76.0 & 0.70 & 1.40 \\
    ViT-S/16 & 50\% & 77.1 & 0.69 & 1.40 \\
    ViT-S/16 & 75\% & 78.6 & 0.69 & 1.39 \\
    ViT-S/8 & 0\% & 79.9 & 0.66 & 1.40 \\
    ViT-S/8 & 50\% & 81.6 & 0.64 & 1.39 \\
    ViT-S/8 & 75\% & 82.1 & 0.64 & 1.38 \\
    ViT-B/8 & 0\% & 80.5 & 0.65 & 1.38 \\
    ViT-B/8 & 50\% & 81.9 & 0.64 & 1.37 \\
    ViT-B/8 & 75\% & 83.6 & 0.63 & 1.37 \\
    \bottomrule
  \end{tabular}
  \caption{Summary of k-NN performance on the three downstream tasks across our experiments. For each experiment, we show the highest accuracy reached during training, regardless of epoch. For experiments that involve pruning, this is usually around 200 epochs, but for the no-pruning experiments, the highest accuracy is often reached earlier during training (Figure \ref{fig:knn_accuracy}), after which the accuracy plateaus, or decreases slightly.}
  \label{tab:impact_pruning}
\end{table}

Dataset pruning also substantially reduces the computational cost required to reach a given level of model performance for TenGeoP classification. For instance, the best TenGeoP classification accuracy for a ViT-S/16 is 76\% when no pruning is applied during training, and it is reached after about 145 epochs. By contrast, the same accuracy is reached as early as after 85 epochs under the 75\% pruning scenario, a more than 40\% gain in computational resources.

\subsubsection{Comparison to other SAR foundation models}

Table \ref{tab:comparison_performance_knn} compares the performance of existing SAR Sentinel-1 foundation models with ours. The performance tends to vary significantly across different downstream tasks, as the relevant information is embedded at different levels within the image. For example, class differences in TenGeoP are strongly linked to global-level features, whereas wind field estimation depends on the finer texture details of the image.

\begin{table}[h!]
  \centering
  \begin{tabular}{@{}lcccc@{}}
    \toprule
    Method & Backbone & TenGeoP & SWH & Wind \\
    \midrule
    CROMA & ViT-B/8 & 65.4 & 0.78 & 1.95 \\
    MoCo & ResNet50 & 60.9 & 0.83 & 1.98 \\
    DeCUR & ResNet50 & 58.3 & 0.84 & 2.10 \\
    DOFA & ViT-B/16 & 58.4 & 0.96 & 2.40 \\
    DOFA & ViT-L/16 & 63.4 & 0.95 & 2.43 \\
    SoftCon & ViT-S/14 & 73.2 & 0.98 & 2.08 \\
    SoftCon & ViT-B/14 & 74.8 & 1.01 & 2.08 \\
    \midrule
    OceanSAR-1 & ResNet50 & 75.5 & 0.75 & 1.62 \\
    OceanSAR-1 & ViT-S/16 & 78.6 & 0.69 & 1.39 \\
    OceanSAR-1 & ViT-S/8 & 82.1 & 0.64 & 1.38 \\
    OceanSAR-1 & ViT-B/8 & \textbf{83.6} & \textbf{0.63} & \textbf{1.37} \\
    \bottomrule
  \end{tabular}
  \caption{Comparison of k-NN performance of existing models and ours across the three selected downstream tasks: (1) classification accuracy on TenGeoP, (2) regression RMSE for significant wave height prediction, and (3) regression RMSE for wind speed prediction.}
  \label{tab:comparison_performance_knn}
\end{table}

Most self-supervised models developed for SAR imagery have been trained on Ground Range Detected (GRD) images in Interferometric Wide (IW) swath mode, which are available in dual polarization (VV+VH). As a result, these models expect two-channel images as input. However, WV images are only available in single-polarization (VV). In Table \ref{tab:comparison_performance_knn}, the results for WV images are obtained by simply duplicating the single-channel VV input to match the expected dual-channel format.

For further evaluation, we also perform supervised fine-tuning for existing SAR SSL models. A wide range of different methods were designed to perform Parameter Efficient Fine-Tuning (PEFT) on pre-trained backbones, like visual fine-tuning \cite{jia2022visual} or adapters \cite{chen2022adaptformer,steitz2024adapters}. Most methods specifically aim at improving the trade-off between the computational cost and performance degradation in fine-tuning.
Note, however, that most of the proposed fine-tuning methods are supervised: in particular, they do not lead to a fine-tuned module that is transferable across downstream tasks. This goes against the spirit of foundation models, and would severely limit the applicability of such fine-tuning approaches for downstream tasks with little annotated data.

Some recent studies developed PEFT methods specific to geospatial foundation models \cite{hu2024tea,selvam2024rapid}, and started to explore the challenge of self-supervised fine-tuning \cite{khannaexplora,scheibenreif2024parameter}.
Others, by contrast, developed supervised fine-tuning methods to address the specific case of cross-domain adaptation by incorporating domain inductive bias for specific cases, e.g. for multi-spectral \cite{2503.09493}, thermal \cite{wang2025cross}, or RGB-Depth \cite{hoffman2016cross} imagery.

As it is not the scope of this paper to design a radar-specific adaptation module, we proceed as follows.
For ResNet backbones like MoCo and DeCUR, we perform full fine-tuning.
For SoftCon, the best performing ViT-based model, we perform a classical LoRA fine-tuning on the patch embedding and attention layers.
Additionally, in each case, we use a linear layer as the classification head to probe model performance. This head is trained jointly with the backbone adapters, using different learning rates. We then compare these results to our self-supervised model under the same evaluation protocol. It is important to note that this setting places our model at a disadvantage, as only the head parameters are trainable—resulting in significantly fewer parameters available to learn the task-specific objective.\\

The fine-tuning results are shown in Table \ref{tab:comparison_performance_knn_tengeop_adapter}. Accuracies on TenGeoP are improved by 10-25\% for existing foundation models and by 5-10\% for our models. Similarly, the RMSE for wave height and wind speed decreases by 0.05-0.2 for existing models and by 0.1-0.3 for our models.

Our specialized foundation models consistently outperform existing foundation models on the three SAR WV tasks, and the gap is especially large for the wave height and wind speed regression (RMSE difference of 0.2-0.6). This can be attributed, in part, to the fact that existing models were trained on different data, which leads to both sensor and domain shifts. Notably, our training dataset consists exclusively of oceanic SAR SLC (single-look complex) images, whereas existing models have been predominantly trained on SAR GRD (ground range detected) imagery over land. Future work could investigate the relative impact of sensor versus domain shift in driving this performance gap.

The smaller performance gaps between models on the TenGeoP benchmark dataset suggest that TenGeoP may be relatively simplistic as a benchmark and may lack sufficient discriminative power to effectively evaluate the performance of self-supervised learning models for SAR imagery. Notably, even with a simple linear probing head—i.e., solving a convex problem—we achieve results that equal the current state-of-the-art in supervised learning on this task \cite{WANG2019111457}.

For our models, we find that the best-performing one on TenGeoP is the ViT-S/16, closely followed by ViT-B/8. However, for the wave height and wind speed regression tasks, larger models with smaller patch sizes perform better. This likely results from TenGeoP classes being primarily characterized by global image features, and being therefore less sensitive to small-scale structures that are better captured by models with smaller patch sizes. By contrast, the other two tasks rely more heavily on local texture information to accurately estimate geophysical parameters.\\

\begin{table}[h!]
  \centering
  \begin{tabular}{@{}lccccc@{}}
    \toprule
    Method & Backbone & TenGeoP & SWH & Wind\\
    \midrule
    MoCo-FT & ResNet50 & 86.5 & 0.77 & 1.80 \\
    DeCUR-FT & ResNet50 & 81.9 & 0.82 & 1.93 \\
    Softcon-LoRA & ViT-S/14 & 84.2 & 0.78 & 1.98 \\
    Softcon-LoRA & ViT-B/14 & 85.1 & 0.79 & 1.95 \\
    \midrule
    OceanSAR-1 & ResNet50 & 80.9 & 0.63 & 1.33 \\
    OceanSAR-1 & ViT-S/16 & \textbf{89.0} & 0.57 & 1.34 \\
    OceanSAR-1 & ViT-S/8 & 85.8 & 0.55 & 1.30 \\
    OceanSAR-1 & ViT-B/8 & 88.3 & \textbf{0.54} & \textbf{1.29} \\
    \bottomrule
  \end{tabular}
  \caption{Linear probing accuracy of existing models (adapted to the single WV polarization or fine-tuned with LoRA) and ours on the three selected downstream tasks.}\label{tab:comparison_performance_knn_tengeop_adapter}
\end{table}

\section{Discussion and Conclusion}

In this study, we introduced a dynamic pruning methodology for self-supervised training of an EO foundation model on Sentinel-1 WV imagery. Given the high redundancy inherent in the dataset, our pruning strategy significantly accelerates convergence.

Beyond efficiency gains, we find that dynamic pruning also improves downstream performance. We attribute this to the regularizing effect of resampling from a redundant dataset, which helps avoid overfitting. From an optimization standpoint, this insight suggests promising connections to classical strategies—such as restart-based methods—that have seen limited exploration in this setting \cite{renegar2022simple,pokutta2020restarting}.

We also address a gap in existing EO foundation models by focusing on a modality—Sentinel-1 WV—for which no pretrained model is well-suited. To support further research, we release a benchmark framework tailored to this modality, relevant for applications ranging from wind field estimation to oil slick detection. We also release the weights for \textit{OceanSAR-1}, the first in a family of models designed specifically for ocean observation using SAR imagery.

Finally, our findings emphasize the complexity of satellite data, where varying acquisition modes result in divergent data distributions. We show that models trained across multiple modes do not necessarily generalize to unseen ones. In fact, for specialized modalities, unimodal self-supervised models can outperform multimodal counterparts while requiring significantly fewer computational resources—further amplified by our pruning strategy.

\section{Acknowledgements}
This work was granted access to the HPC resources of IDRIS and TGCC under the allocation 2025-[A0171015666] made by GENCI. We also thank Vivien Cabannes for valuable discussions and insights.

\section{Data and Code}
The OceanSAR-1 model weights are available on HuggingFace at \href{https://huggingface.co/galeio-research/OceanSAR-1}{https://huggingface.co/galeio-research/OceanSAR-1}. We also release the probing weights for the TenGeoP classification task and the wave height and wind speed regression tasks.

\newpage

{
    \small
    \bibliographystyle{ieeenat_fullname}
    \bibliography{biblio}
}

\end{document}